\documentclass[runningheads]{llncs}
\usepackage{graphicx}
\usepackage{amsmath,amssymb} % define this before the line numbering.
\usepackage{color}
\usepackage{graphicx}
\usepackage{graphicx}
\usepackage{tikz}
\usepackage{comment}
\usepackage{amsmath,amssymb} % define this before the line numbering.
\usepackage{color}
\usepackage{float}
\usepackage[pagebackref,breaklinks,colorlinks]{hyperref}
\usepackage{subfig}
\usepackage{booktabs}
\usepackage{epstopdf}
\usepackage{arydshln}
\usepackage{multirow}
\usepackage{marvosym}
\begin{document}
\pagestyle{headings}
\mainmatter

\def\ACCV22SubNumber{200}  % Insert your submission number here

%===========================================================
\title{$M^2$-Net: Multi-stages Specular Highlight Detection and Removal in Multi-scenes} % Replace with your title
\titlerunning{Multi-stages Specular Highlight Detection and Removal in Multi-scenes}
\authorrunning{Z. Huang et al.}

\author{Zhaoyangfan Huang, Kun Hu, Xingjun Wang\Letter}
\institute{Tsinghua Shenzhen International Gradual School, Shenzhen, China \email{\{hcyf20,Wang.Xingjun\}@mails.tsinghua.edu.cn}}

\maketitle

\vspace{-0.5cm}
\begin{figure}[h]
	\centering
	\includegraphics[scale=0.5]{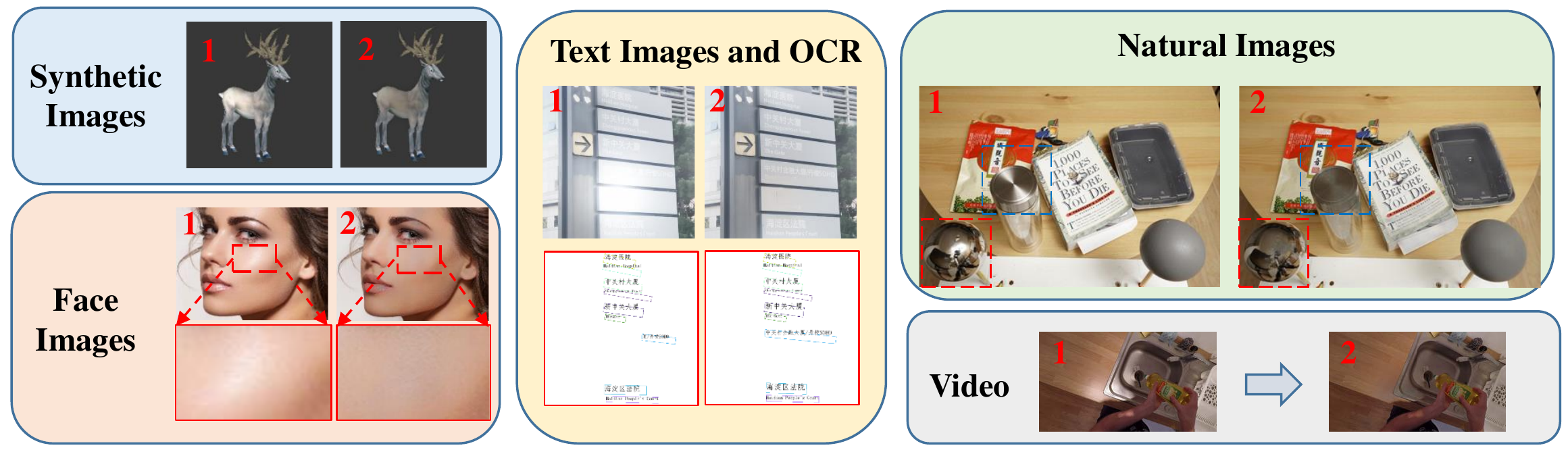}
	\caption{ The results of this paper show the comparison between the original highlight images and the removed highlight images for a variety of scenes (including synthetic images, face images, text images and natural images) for the network. The \textbf{\color{red} 1} in the upper left corner of the image represents the original highlight image, and the \textbf{\color{red} 2}  represents the image with the highlights removed using the network architecture of this paper. There are also zoomed-in comparison images, text OCR results shown in the scenes and video highlight removal result.}
% 	本文的网络在多种场景（包括传统算法使用的合成图像、人脸图像、文字图像和自然图像）下原始高光图像与去除高光图像的对比结果展示。图像中左上角的1代表原始高光图像，2代表使用本文网络架构去除高光的图像。本文场景下还有放大的对比图和后续文字识别结果的展示。
	\label{allinone}
\end{figure}
\vspace{-0.8cm}

\begin{abstract}
% 本文中，我们创新性地提出了一种端到端统一性的高光去除框架。该框架能处理多场景，包括人脸、实物和肠胃镜图像，并且能得到超越最先进算法的性能。本文框架主要有三个部分组成，高光特征图分离模块、高光粗略去除模块和高光精细去除模块。首先高光去除模块能直接从原始带有高光的图像中分离出高光图像和非高光图像。然后使用基于上下文特征提取的高光粗略分离网络得到去除高光的图像。为了进一步提高高光去除的效果，最后使用精细高光去除网络得到精细化的高光去除图像。在多场景下大量实验结果表面本文的框架能获得卓越的高光去除视觉效果，并且在多个量化评价指标上都取得了最先进的结果。
In this paper, we propose a novel uniformity framework for highlight detection and removal in multi-scenes, including synthetic images, face images, natural images, and text images. The framework consists of three main components, highlight feature extractor module, highlight coarse removal module, and highlight refine removal module. Firstly, the highlight feature extractor module can directly separate the highlight feature and non-highlight feature from the original highlight image. Then highlight removal image is obtained using a coarse highlight removal network. To further improve the highlight removal effect, the refined highlight removal image is finally obtained using refine highlight removal module based on contextual highlight attention mechanisms. Extensive experimental results in multiple scenes indicate that the proposed framework can obtain excellent visual effects of highlight removal and achieve state-of-the-art results in several quantitative evaluation metrics.  Our algorithm is applied for the first time in video highlight removal with promising results. Our code is available at: \url{https://github.com/hzzzyf/specular-removal}
\end{abstract}

%===========================================================
\section{Introduction}
\label{sec:intro}

Specular highlight is a common phenomenon in the real world, and it has been a long-standing problem in computer vision and image processing. An image taken by a camera usually exhibits specular highlights caused by the shiny material surface of the object when illuminated. Specular highlight appear frequently in many scenes, such as objects surface~\cite{fu2021multi,wu2020deep,fu2020learning} and human faces~\cite{muhammad2020spec}, which affects the visual effect while posing great challenges to computer vision tasks, such as image segmentation~\cite{kim2013specular}, edge detection~\cite{feris2004specular}, text recognition~\cite{wang2020decoupled,hou2021text}, etc. The previous algorithm can only produce satisfactory highlight removal results for one of the scenes, but not uniformly for all scenes. The algorithm proposed in this paper can handle highlight removal in multiple scenes simultaneously and can enhance subsequent computer vision tasks, such as optical character recognition (OCR). The highlight removal effect cases of the proposed method is shown in Fig.~\ref{allinone} for different scenes and video. 
% 镜面高光频繁出现在玻璃物体表面、人脸和肠胃内镜中，影响视觉效果的同时给计算机视觉任务带来极大的挑战，例如图像分割。之前的算法只能针对其中一种场景做出令人满意的高光去除结果，并没不能针对所有场景的高光图像进行统一的去除。 本文提出的算法能同时处理多种场景下的高光去除，并且能提升后续的计算机视觉任务的效果，例如光学文字识别。图1中展示了不同场景下本文提出的算法的高光去除效果。
% \textbf{\color{red}Therefore, removing specular highlights facilitates the improvement of image quality, thus enhancing the effectiveness of computer vision tasks, such as image segmentation~\cite{kim2013specular}, edge detection~\cite{feris2004specular}, text recognition~\cite{wang2020decoupled}, steganography~\cite{tancik2020stegastamp}, etc.}

\begin{figure}[h]
	\centering
	\includegraphics[scale=0.38]{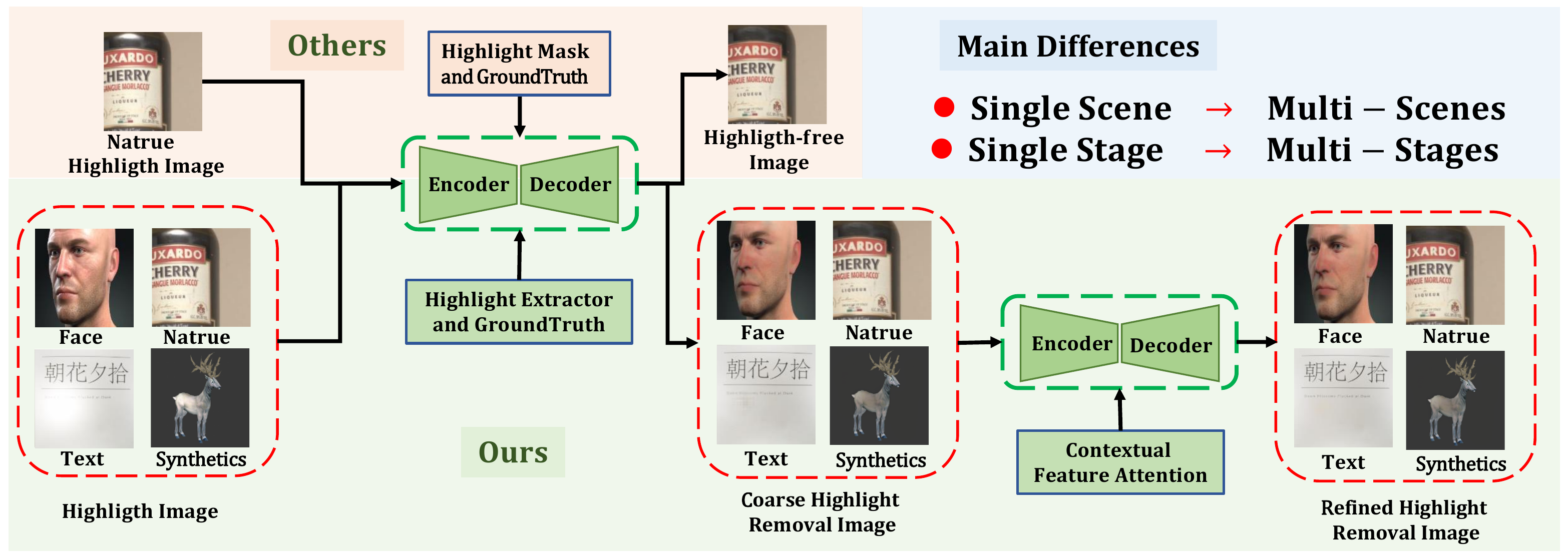}
	\caption{The algorithm proposed in this paper differs from the existing algorithms. The previous algorithms for pink background can only process images for specific scenes, such as natural images, face images, etc. While our algorithm can handle highlight removal for multiple scenes at the same time. And our algorithm for green background adds a multi-stage refinement highlight removal network to the previous single-stage algorithm. This ensures that the obtained highlight removal images are closer to the real scenes.}
	\label{maindifferences}
\end{figure}
% Separating the specular highlight from a single image is an ill-posed problem~\cite{guo2014robust}. 
% Therefore, more additional constraints are required.
% In academics, there are several methods for removing specular highlights.
Early works typically remove highlight based on different constraints or assumptions, such as using colour reflection model~\cite{huang2016double}, dichromatic reﬂection model~\cite{shafer1985using},  assuming special~\cite{tan2006separation,koirala2009highlight}, and adopting optimization~\cite{kim2013specular,liu2015saturation,fu2019specular}.
% However, all of the preceding methods meet some limitations and may fail to handle some real-world highlight images since their methods can not naturally separate the highlight layer and restore the contents of it.
% The generality of the constraints or assumptions determines the performance of the models.
Recently, deep learning approaches have emerged in the field of specular highlight removal and have achieved remarkable improvements.
% Yi et al.~\cite{yi2020leveraging} proposed an unsupervised method for separating highlight by leveraging multi-view image sets.
% Zhu et al.~\cite{zhu2020highlight} adopt the structure of conditional generative adversarial network (CGAN) to generate highlight-free facial images.
% Madessa et al.~\cite{madessa2020deep} proposed a partial convolution-based inpainting method for specular highlight removal from transmissive materials.
% Despite the objective progress made by learning-based methods, these methods suffer from two weaknesses.
% First, deep learning methods heavily rely on training data, but some methods like~\cite{madessa2020deep,tan2005reflection,tan2008separating} using synthetic data may fail to remove the highlight in the real-world images.
% Second, some of their methods rely on the highlight mask as the label to train the network.
However, both traditional algorithms and deep learning-based algorithms only use a single-stage algorithm and remove highlight in a single scene, resulting in unsatisfactory results in other scenes. The comparison in Fig.~\ref{maindifferences} demonstrates that the two most significant differences between the algorithm in this paper and other existing algorithms are multi-scenes and multi-stages.

% Both traditional and deep learning methods only remove highlights in a single scene, which makes them unsatisfactory when applied to other scenes.

% 图2对比展示了本文算法与现有其他算法的最主要的两个不同点是多场景和多阶段。
% these highlight masks labeled by hand or algorithm algorithm may not effectively capture the highlight information, thus resulting in performance deteriorations on the final results.
% 包含提取高光区域的高光特征提取模块，粗略高光去除模块，运用上下文注意力机制的精细化高光去除模块。具体的，高光检测模块使用多尺度信息提取具有不同强度高光区域。粗略去除高光模块能去除基础的高光。受到图像填补的启发，我们使用具有上下文注意力机制的精细化高光去除模型对粗略高光结果进行进一步的效果提升。

In this paper, we propose a novel highlight removal model for multi-scenes specular highlight removal. 
It includes a highlight feature extraction (HFE) module to extract highlight regions, a coarse highlight removal module, and a refine highlight removal module with a contextual feature attention (CHA) mechanism. Specifically, the highlight extraction module uses multi-scale information to extract regions with different intensity highlights. The coarse highlight removal module removes the underlying highlights with Gated Convolutions and Dilated Convolutions~\cite{yu2019free}. Inspired by image inpainting~\cite{chang2019free}, we use a refine highlight removal model with a contextual attention mechanism to further improve the effect of the coarse highlight results.
% It contains a highlight feature extractor (HFE) module and a contextual highlight attention (CHA) mechanism. 
% Specifically, the HFE module is proposed to search for the highlight region and extract the highlight features from a single image, which can capture the information between highlight region and non-highlight region.
% The CHA mechanism is designed to obtain the contextual feature from the highlight images, which enables the removal of highlight and restoration of the images with limited information. 
% Moreover, inspired by the idea of image inpainting~\cite{chang2019free}, a generative adversarial network is adopted as the framework of our network, and the Gated Convolutions and Dilated Convolutions~\cite{yu2019free} is employed to improve the removal quality and speed.
% Using only paired highlight and non-highlight images and no additional highlight mask or highlight intensity, 
Our method achieves highlight removal with satisfactory quality in most real-world situations and outperforms existing learning-based techniques. 
To summarize, three main contributions of this paper as follows:
\begin{itemize}
	\item We propose a unified multi-scene highlight removal framework capable of handling synthetic images, face images, text images and natural images.
% 	我们提出了统一性的多场景高光去除框架，能处理人脸图像、自然图像和肠胃镜图像。
	\item We propose a highlight feature extractor module and a contextual feature attention mechanism, which can effectively detect highlight locations and perform highlight removal. Moreover, two-stage (coarse and refine stages) highlight removal algorithm makes highlight removal more satisfying. 
% 	我们提出了高光检测模块（特征金字塔）和上下文注意力机制，能有效的检测高光位置并进行高光的去除。
	\item Experimental results on multiple image datasets outperform existing state-of-the-art algorithms and perform very well in video highlight removal. 
% 	多个图像数据集上的实验结果均能超越现有的最先进算法，并且在视频高光去除中表现也非常出色。

\end{itemize}
% The rest of the paper is organized as follows. Section~\ref{RelatedWork} provides
% an overview of related work. In Section 3, we discuss Spec-Face.
% Section 4 explains our methodology for training and testing of Spec-
% Net and Spec-CGAN. In Section 5, we describe an experimental
% evaluation, and in Section 6, we conclude and discuss future work.
%Do not use any additional Latex macros.

\section{Related Work}

\label{RelatedWork}

%-------------------------------------------------------------------------
\noindent\textbf{Highlight detection}. 
Under the assumption that the highlight area is only small, the most used method in highlight detection is the threshold detection method~\cite{park2003truncated}. Although threshold-based highlight detection algorithms are efficient in detecting highlight areas, most of them are very sensitive to the threshold value~\cite{ortiz2005comparative,li2019specular}.  Yang et al.~\cite{yang2014efficient} treat highlight pixels as noise and use a low-pass bilateral filter to smooth out the highlights.
By using multi-scale context contrasted features, Fu et al.~\cite{fu2020learning} create a deep learning-based specular highlight detection network.
% Deep learning-based highlight detection algorithms can effectively exclude non-highlight regions. 
There have been encouraging advances in highlight detection, but they are all algorithms for individual scenes, and there are no highlight detection algorithms for all scenes.

\noindent\textbf{Highlight removal}.
% Single-image reflection removal methods are the most common, including both traditional and learning-based methods.
Highlight removal algorithms can be classified as single-image based and multi-image based. In this paper, we focus on highlight removal for single images, and the highlight removal algorithm for multiple images can be referred to ~\cite{lee1992detection,wang2016light,lei2021robust}.
Tan et al.~\cite{tan2004separating} separated the two reflection components based on the distribution of specular and diffuse points in a two-dimensional maximum chromaticity-intensity space.
Yang et al.~\cite{yang2013separating} utilized the HSI color space to separate diffuse and specular reflection components for color images and proposed an approach to adjust saturation of specular pixels to the values of diffuse-only pixels with the same diffuse chromaticity.
Some methods are to derive a synthetic diffuse image that exhibits the specular highlight~\cite{tan2005reflection,tan2008separating}.
Yang et al.~\cite{yang2010real} presented an effective real-time specular highlight removal method by utilizing the bilateral filtering.
A sparse and low-rank reflection model has been proposed for highlight removal and detection~\cite{guo2018single}, by regarding the task as a nuclear norm and $l_1$-norm minimization problem been solved by Lagrange multiplier method.
Recently, deep learning-based methods are emerging in the field of highlight removal.
Muhammad et al.~\cite{muhammad2020spec} designed two network models (Spec-Net and Spec-CGAN), which are utilized for removing specular highlight of facial Images.
% The method proposed by Funke et al.~\cite{funke2018generative} leveraged the generative adversarial networks for specular highlight removal in endoscopic images, in which they trained a residual convolutional neural network (CNN) to detect and remove specular highlights with weakly labeled data and employed a GAN to assess the results of the CNN during training.
All of the above methods have achieved remarkable performance in highlight removal.
However, they are difficult to implement in large-scale highlight regions, and they are unable to restore highlight images with colored illuminations or complicated textures.

% \noindent\textbf{Multiple-image-based highlight removal}.
% % To handle the ill-posed problem of highlight removal and produce high-quality images, multiple-image-based approaches have been proposed~\cite{guo2014robust}.
% Several researchers used images with varied viewing angles to remove highlight and got promising results.
% Lee et al.~\cite{lee1992detection} leveraged the multiple color images from different viewing directions to detect the specular highlight.
% By utilizing the light field images, Wang et al.~\cite{wang2016light} removes specularity and improves image quality with a color variance analysis of multiple views.
% Guo et al.~\cite{guo2014robust} proposed the RPCA method which removes highlight by leveraging the video/image sequence.
% Lei et al.~\cite{lei2021robust} present a method in which they utilize the reflection-free from a pair of flash and no-flash images and effectively achieve reflection removal.
% Although their methods have proved their effectiveness, the multiple images may not be as accessible.

\noindent\textbf{Image inpainting}.
Unlike highlight removal, the purpose of image inpainting is to fill in the missing areas of images with reasonable content.
The methods proposed by
% \cite{ballester2001filling,wexler2007space,efros2001image,barnes2009patchmatch} 
~\cite{barnes2009patchmatch} typically complete the missing part by leveraging patch similarity and transferring the contents from the background of the image to the missing region.
Recently, deep learning-based approaches have obtained significant attention
owing to the capabilities of reasoning and extracting semantic information.
CNN-based methods have been designed for image inpainting and achieve remarkable
% ~\cite{pathak2016context,song2018contextual,iizuka2017globally,kohler2014mask}.
performance~\cite{song2018contextual}.
Yu et al.~\cite{yu2018generative} proposed the contextual attention and gated convolution~\cite{yu2019free} to acquire the textual of inpainting images, which led to significant progress in image inpainting.
When employing these methods for highlight removal, a highlight mask is required to remove hightlight regions that might otherwise result in the loss of textual content.
Therefore, these methods are not applicable for highlight removal.

\begin{figure*}[ht]
	\centering
	\includegraphics[scale=0.6]{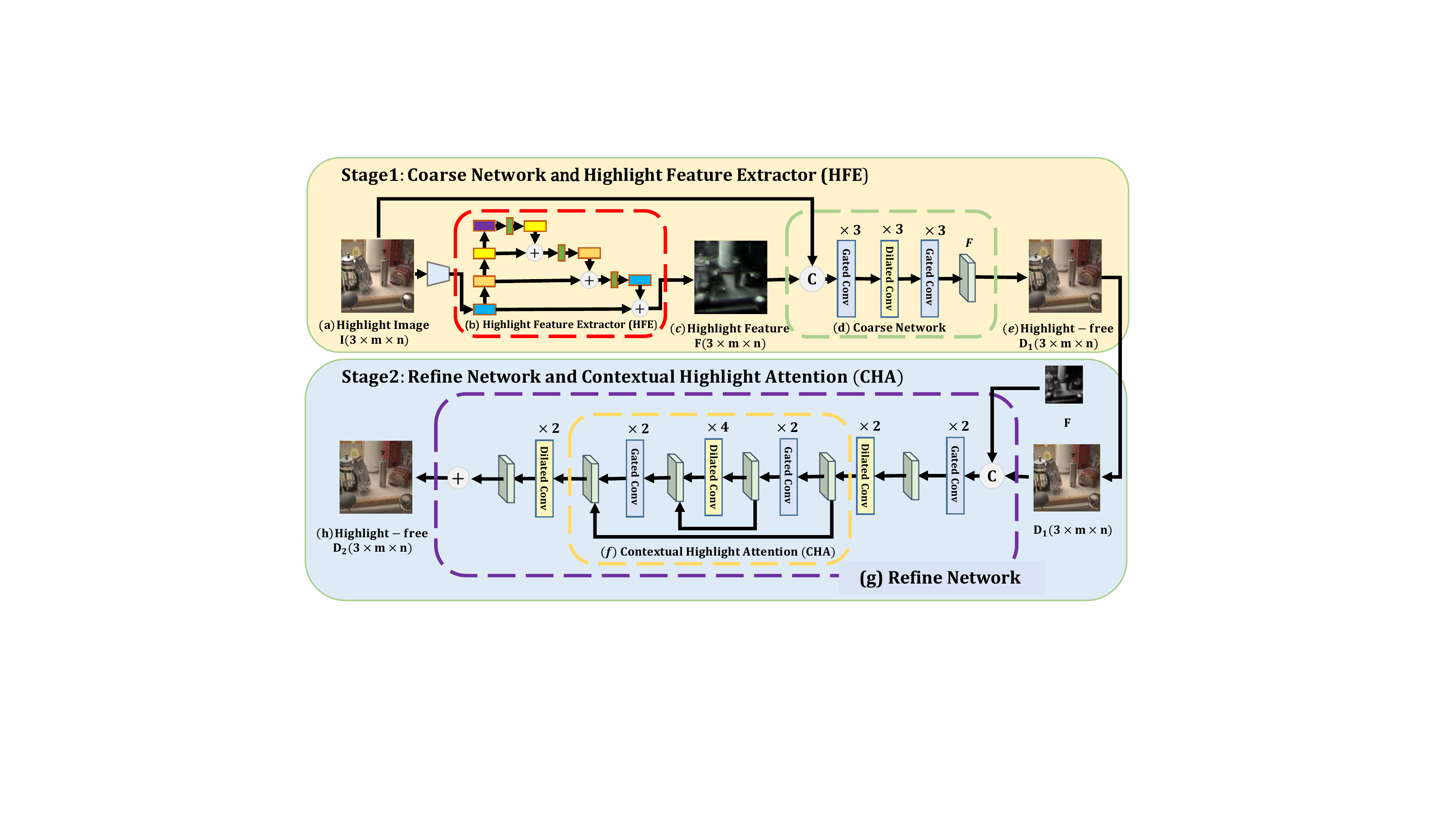}
	\caption{ An overview of the proposed network. Stage 1 includes highlight feature detection network and coarse highlight removal network. Stage 2 includes contextual highlight attention and refined highlight removal network. (a) highlight image $I(3\times m \times n)$; (b) Highligth feature extractor; (c) Highligth feature $F(3\times m \times n)$ of the highlight location obtained by the highlight detection extractor module (HFE); (d) Coarse highlight removal network; (e) Highlight-free image $D_1(3\times m \times n)$ obtained using coarse removal network; (f) Contextual highlight attention (CHA) module; (g) Refinement highlight removal of the coarse highlight removal image $D_1(3\times m \times n)$ and the highlight feature image $F(3\times m \times n)$; (h) Highlight-free image $D_2(3\times m \times n)$.
% 	Firstly, given a input highlight image $I$, the proposed HFE module produces the highlight feature $T$ which is three-channel. Then, the coarse network generates the coarse image $A_{coarse}$. Meanwhile, the contextual highlight attention (CHA) module will compute the attention scores and superimpose them on subsequent images. After, the refine network combines the attention scores and features from the coarse image to produce the refine image $A_{refine}$. Finally, the network output the final non-highlight image.
}
	\label{network}
\end{figure*}
\section{Proposed Method}

\subsection{Motivation}
Fu et al.~\cite{fu2020learning} observed that the intensity of highlights in natural images is high, but the distribution is sparse. Based on the above observations, a multi-scale contextual contrast feature module is proposed to detect highlights. In this paper, we are inspired to design a highlight removal model based on multiscale contextual features.

In image inpainting, a network structure with coarse extraction and refined extraction is used by Yu et al.~\cite{yu2019free}. In this paper, we modify the structure based on this and use it for inpainting after highlight detection.
\begin{figure*}[ht]
	\centering
	\includegraphics[scale=0.5]{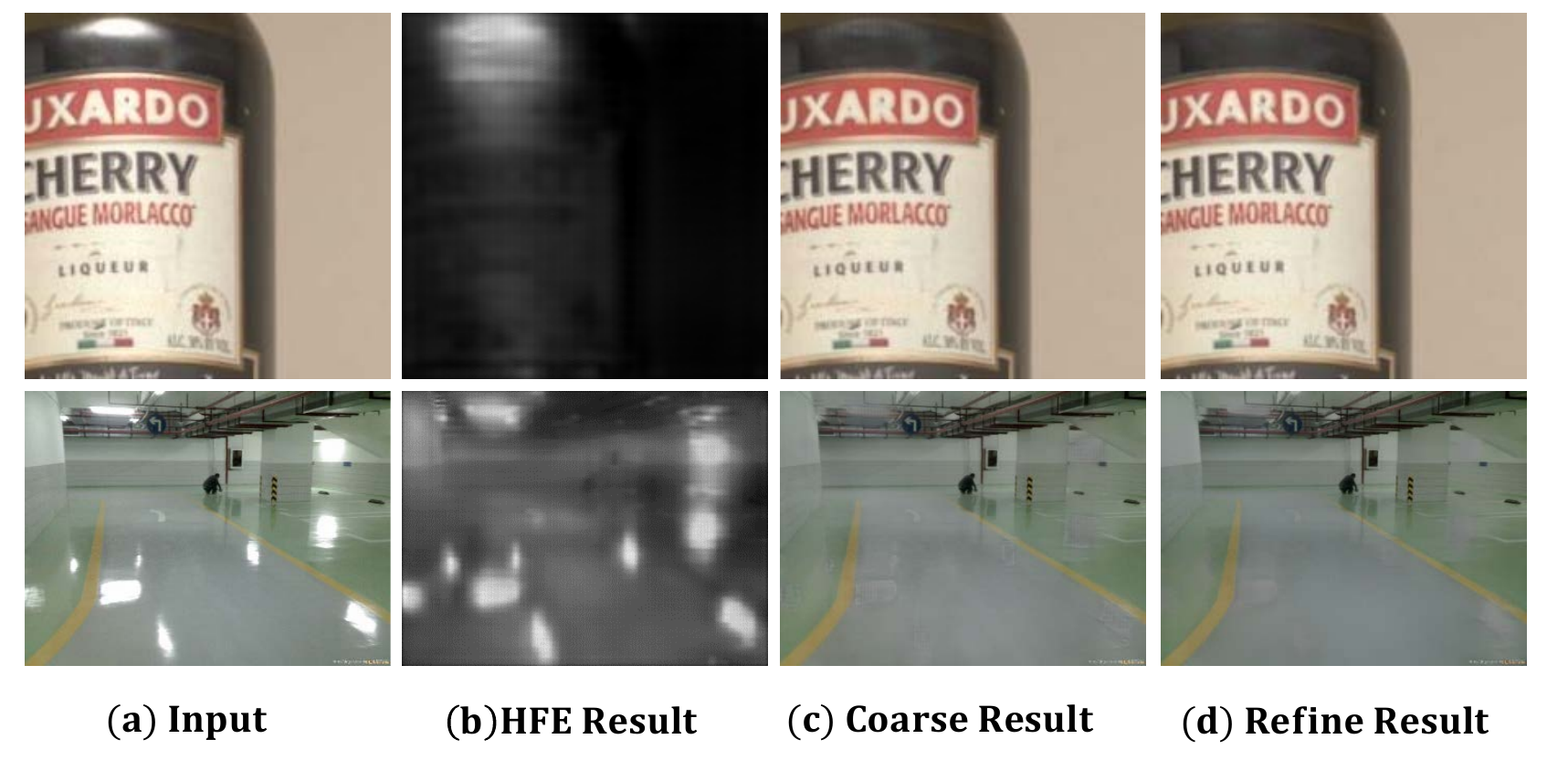}
	\caption{Images of the algorithmic process results in this paper. (a) Input highlight image; (b) Extracted highlight feature images; (c) Coarse highlight removal images; (d) Refine the image for highlight removal. }
	\label{Stage1and2}
\end{figure*}
\subsection{Highlight Feature Extractor Module (HFE)}\label{HFE}
% As shown in Figure \ref{network}, we designed a HFE module to produce a highlight feature as a coefficient matrix to separate the highlight layer from a single image.
% Figure \ref{HIE} illustrates the structure of the HFE module.
% It can be generally regarded as a type of encoder-decoder.
Given a highlight image $I(x, y)$ with size of $m \times n$,
we first adopt a series of resnet~\cite{he2016deep} to acquire $U$ layers of $N$ feature maps $F^{u}_i(x, y), i=1,2,...,N, u=1,2,...,U$ with size of $m/(2^{u-1}) \times n/(2^{u-1})$. In this paper, we set $N=3, U=4$.
For each $v=1,2,...,N-1$, we consider a transposed convolutional layer with $M$ of filters $W_j(x, y), j=1,2,...,M$ to obtain $UpF^{v}_i(x, y)=\sum_{j=1}^{M} W_{j}(x, y)\cdot F^{v+1}_i(x, y), i=1,2,...,N, v=2,...,U$. 
For each $v=1,2,...,N-1$, $K$ of filters $W_{k}(x, y), k=1,2,...,K$ are made to generate the highlight feature by using $HF_{i}(x, y)= \sum_{k=1}^{K} W_{k}(x, y)\cdot (F^{v}_i(x, y)  + UpF^{v+1}_i(x, y))$.
% \begin{equation}
%     HF_{i}(x, y)= \sum_{k=1}^{K} W_{k}(x, y)\cdot (F^{v}_i(x, y)  + UpF^{v+1}_i(x, y)).
%     \label{eq:HFE}
% \end{equation}
% 高光提取模块的作用就是检测高光与非高光区域。与现有算法检测高光区域的0-1结果不同，我们使用强度系数代表属于高光区域的可能性。系数越大，代表高光区域的可能性越大。
The role of the highlight extraction module is to detect highlight and non-highlight regions. Unlike existing highlight detection algorithms that detect highlight regions with 0 or 1 results, we use intensity coefficients to represent the likelihood of belonging to highlight regions, shown in Fig.~\ref{network} (c). 
% The larger the coefficient, the more likely it represents the highlight region.
The larger the coefficient, the greater the intensity of the highlights
Different from the existing single-channel highlight masks, the highlight feature extraction module proposed in this paper can extract highlight features from each channel, which means it can better handle real-world highlights. This is because real-world highlights behave differently in the three-channel, rather than simply as a result of single-channel reflections.
Fig.~\ref{Stage1and2} (b) shows the results of highlight extraction. It is clear that the higher the highlight intensity, the higher the highlight feature coefficients. And the extraction effect is obvious, most of the highlight regions can be extracted.

\subsection{Coarse Network and Refine Network}\label{CN}

Architecture designs based on coarse and refined networks are widely used in various computer vision tasks. The network structure has the obvious advantage of being able to balance the whole and the details, such as 3D point cloud completion~\cite{wang2020cascaded}, rain removal~\cite{wang2021attentive}, shadows removal~\cite{chen2021canet},
semantic segmentation~\cite{ding2021scarf}. Since the highlight removal task in this paper has many similarities with the shadow removal task, the network structure of this paper has been slightly modified from~\cite{chen2021canet}.
% 图4中的b展示了高光粗略去除和高光精细化去除的结果。粗略去除的效果主要将高光区域进行了基础的去除，高光填补的效果不自然。通过精细化去除网络的处理，高光填补的像素与周围的像素更加贴近，更加自然。
Fig.~\ref{Stage1and2} (c) and (d) shows the results of coarse highlight removal and refine highlight removal. The effect of coarse removal mainly removes the highlight region in a base-level, and the highlight-filled effect is unnatural. By refining the removal network, the highlight-filled pixels are more closely matched to the surrounding pixels and more natural.
% 基于粗略网络和精细化网络的结构设计在各种计算机视觉任务中得到广泛应用。该网络结构的优势很明显，能兼顾整体和细节。例如Lu等人将粗略-精细化网络用于3D点云的生成任务上、Hu等人在去雨网络中使用该结构得到很好的效果。 由于本文的高光去除任务与阴影去除任务具有很多相似的地方，所以本文的网络结构在论文47的基础上进行了细微的改动。

\subsection{Contextual Highlight Attention Mechanism (CHA)}\label{CHA}

% Our attention mechanism consists of two key components: Contextual Feature Separation Module and Attention Computing Module.

% 上下文高光注意力机制，之前的上下文注意力机制主要是针对利用背景的信息对图像中缺失的部分进行补全，而我们提出的高光上下文注意力机制的将不仅利用背景的信息，同时还利用原高光区域内的剩下的信息来补全高光部分缺失的信息。
% 特别的是，我们将特征分为高光特征与背景特征，并将其进行多尺度的分块，通过获取高光内外的联系度来计算注意力分数。
% 
% In particular, previous learning-based methods of highlight removal usually adopt the convolutional neural layers to obtain the highlight-free features.
% However, it is not effective for borrowing distant features from the images' background.
% In this paper, to handle this problem, we innovatively introduced the contextual highlight attention(CHA) mechanism to transfer the contextual feature from the non-highlight regions to highlight regions.
% \noindent\textbf{Contextual Feature Separation Module}.
% To 
% \begin{equation}\label{eqi}
% 	\begin{aligned}
%     &T_f = conv(Wg,T) \\
%     &h = T_f  \odot F \\
%     &b = (1 - T_f)  \odot F
%     \end{aligned}
% \end{equation}
% \begin{equation}\label{eq10}
% 	\begin{aligned}
% 		\mathcal{L}_{d} = &\mathbb{E}_{x\sim P_{data}(x)}[max(0,1-D(x,T))] + \\
% 		&\mathbb{E}_{z\sim P_{g}(z)}[max(0,1+D(G(z),T))]
% 	\end{aligned}
% \end{equation}
% 为了将高光特征和背景特征从上下文特征中分离出来，我们首先将HFE模块中提取的高光特征通过一个卷积层下采样到与上下文特征相同的维度。让后将
%--------------------------------问题需要讨论的---------------------------
% 高光区域与非高光区域如果划分的？高光区域的填补是否只使用了非高光区域的信息，高光区域自己的信息是如何使用的？
% \noindent\textbf{Compute attention scores}.
\begin{figure*}[ht]
	\centering
	\includegraphics[scale=0.42]{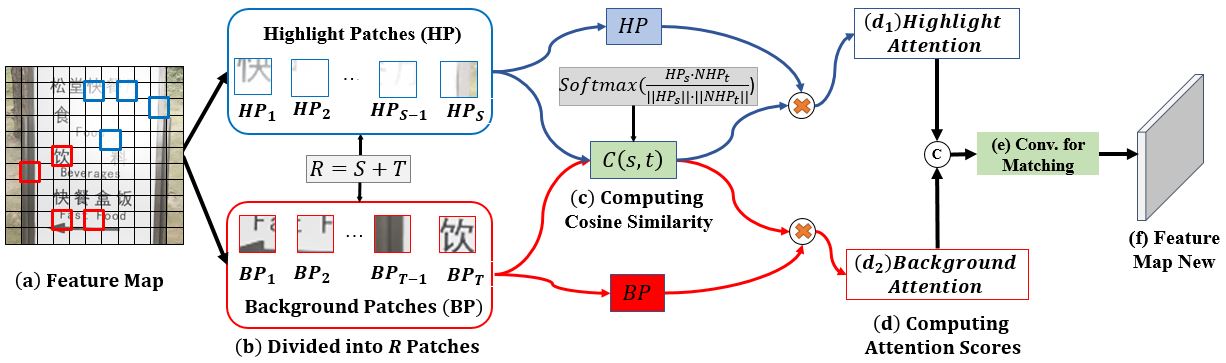}
	\caption{Illustration of our contextual highlight attention mechanism. The highlight feature image in (a) is divided into $s$ highlight patches (HP) and $t$ background patches (BP) (non-highlight patches) according to the result of the highlight feature extraction module, and the total number of patches is $r$ in (b). Then, we calculate the cosine similarity $C(s, t)$ between each of the highlight patches and each of the background patches in (c). Then the corresponding coefficients are multiplied between the highlight and background patches to obtain the highlight attention (HA) in ($d_1$) and background attention (BA) in ($d_2$). After, the two are connected and the patches in the background patches that are similar to the highlight patches are found by the convolutional matching module in (e). Finally we get the output highlight attention feature map in (f).}
% 	按照高光特征提取模块的结果将特征图像分成t个高光块和背景块（非高光块），总块数为r个。计算高光块中每一个与背景块中每一个块之间的余弦相似度c(s, t)。然后将对应系数与高光块与背景块之间进行相乘，得到高光注意力和背景注意力。接着将两者连接起来，通过卷积匹配模块找到背景块中与高光块相近的块。
	\label{cha}
\end{figure*}
The contextual attention was widely used in the image inpainting field~\cite{song2018contextual,yi2020contextual} to generate the missing region from the background information.
While the previous contextual attention mechanism focuses on using the background information to fill in the missing parts of the image, our CHA not only leverage the background contextual features but also the highlight region features.
In particular, we divide the contextual features into highlight features and background features, and piecemeal them at multiple scales shown in Fig.~\ref{cha}. 

For highlight feature image $HF(x, y)$ with size of $m \times n$, we first divide the $HF$ into a number of small patches ($SP_{r}, r=1,2,...,R; R=m/l \times n/l$) of length $l=2$. The $SP_{r}$ is divided into highlight regions ($HP_{s}, i=1,2,...,S, S \in [1,  m/l \times n/l)$) and non-highlight regions ($BP_{t}, t=1,2,...,T, T \in [1,  m/l \times n/l)$) by highlight features $HF$, where $SP_{r} = HP_{s} \cup BP_{t}$. To retrieve the matched patches, we consider calculating the cosine similarity of two different patches to acquire the attention scores   $C(s, t)=softmax(\frac{HP_{s} \cdot BP_{t}} {\|HP_{s}\|\|BP_{t}\|})$.
% \begin{equation}
%     C(s, t)=softmax(\frac{HP_{s} \cdot BP_{t}} {\|HP_{s}\|\|BP_{t}\|}).
%     \label{softmax}
% \end{equation}
 After calculating the attention scores, the attention scores will be applied on the feature map extracted from non-highlight regions to fill the highlight regions. The highlight regions is filled using the non-highlight regions, and the highlight regions after filling are $HA_{s}= \sum_{t=1}^{T}BP_{t} \cdot C(s, t).$
%  \begin{equation}
%       HA_{s}= \sum_{t=1}^{T}BP_{t} \cdot C(s, t).
%       \label{attention1}
%  \end{equation}
 In order to make the image after filling closer to the surrounding pixels, this paper uses non-highlight regions to fill the highlight regions while retaining some pixels of the original highlight regions by using $BA_{t}= \sum_{s=1}^{S}HP_{s} \cdot C(s, t).$ $HA_s$ and $BA_t$ are convolutionally matched to obtain the final output new feature image.

\subsection{Loss Functions}\label{4.4}

For acceptable performance, we design the loss function of our network, which consists of an adversarial loss and a removal loss.
For adversarial loss, we adopt the hinge loss for the discriminator to identify whether the input image is generated.
it is expressed as:

\begin{equation}\label{eq10}
	\begin{aligned}
		\mathcal{L}_{d} = &\mathbb{E}_{x\sim P_{data}(x)}[max(0,1-D(x,HF))] + \\
		&\mathbb{E}_{z\sim P_{g}(z)}[max(0,1+D(G(z),HF))]
	\end{aligned}
\end{equation}
where $HF$ is highlight feature extracted by proposed HFE module.
$D(.)$ denotes the discriminator and $G(.)$ is the generator.
$x$ represents GT images and $z$ is input highlight images.
$P_{data}(x)$ and $P_{g}(z)$ are distribution of $x$ and $z$.

The removal loss contains three parts including a gan loss $\mathcal{L}_{g}$, a content loss $\mathcal{L}_{content}$, and a perceptual loss $\mathcal{L}_{per}$.
The Gan loss is written as:
\begin{equation}\label{eq11}
	\mathcal{L}_{g} = \mathbb{E}_{z\sim P_{g}(z)}[D(G(z),HF)]
\end{equation}

The content loss $L_{content}$ is used to maintain the visual content between the GT images and highlight removal images generated by the proposed network.
\begin{equation}\label{eq12}
	\begin{aligned}
		\mathcal{L}_{content}= &\left \| G_{coarse}(z)-x \right \|_{1}+ 
		\left \| G_{refine}(z)-x \right \|_{1}
	\end{aligned}
\end{equation}
where $G_{coarse}(.)$ and $G_{refine}(.)$ denote the output of coarse network and refine network.

Finally, the perceptual loss plays an important role in ensuring content similarity of the real non-highlight image and generated images.
We employ a pre-trained VGG-16 model\cite{simonyan2014very} to extract low-level feature maps, and then the perceptual loss can be expressed as the ${\ell_{1}}$ norm between feature maps of GT and those highlight removal images.

\begin{equation}\label{eq12}
	\begin{aligned}
		\mathcal{L}_{per}= \left \| \phi (G(z))-\phi (x) \right \|_{1}
	\end{aligned}
\end{equation}
where $\phi(.)$ represents the output of the pre-trained VGG16 network.

The overall removal loss function for training the generator is thus expressed as:

\begin{equation}\label{eq12}
	\begin{aligned}
		\mathcal{L}_{rem}= \lambda _{g}\mathcal{L}_{g}+\lambda_{content}\mathcal{L}_{content}+\lambda_{per}\mathcal{L}_{per}
	\end{aligned}
\end{equation}
where $\lambda _{g}$, $\lambda_{content}$, $\lambda_{per}$ represent the weighs of gan loss, content loss, and perceptual loss respectively. 
In this paper, we experimentally set it to $\lambda_{content}=10$, $\lambda_{per}=1$, $\lambda _{g}=1$.

\section{Experiment}

\subsection{Setup}
% \textbf{Datasets:}
% There are not many published highlight datasets, and most of these datasets are synthetic such as LIME\cite{meka2018lime}.
We use the Specular Highlight Image Quadruples (SHIQ) dataset~\cite{fu2021multi} and SD1~\cite{hou2021text}. SHIQ is a high quality real-world highlight dataset including 10k training quadruples and 1k test quadruples and SD1 is a text highlight images datasets including 12k training data and 2k test data.
% In addition, to make highlight removal more effective in the real world, we proposed a Real-World Highlight Images (RWHI) dataset which consists of 1k pairs of real-word highlight and non-highlight images.
% More details about the dataset are in the supplementary material.
% In general, we use two datasets to train our network, both SHIQ and RWHI, and the remaining portion of them to evaluate our model.
%
%\vspace{-0.5cm}
%
%\begin{figure}[H] 
%	\centering 
%	\begin{minipage}[b]{0.9\linewidth} 
%		\subfloat[Highlight]{
%			\begin{minipage}[b]{0.24\linewidth} 
%				\centering
%				\includegraphics[width=\linewidth]{./pic5/1.png}\vspace{1pt}
%			\end{minipage}
%		}\hspace{-7.5pt}
%		\hfill
%		\subfloat[Non-highlight]{
%			\begin{minipage}[b]{0.24\linewidth}
%				\centering
%				\includegraphics[width=\linewidth]{./pic5/2.png}\vspace{1pt}
%			\end{minipage}
%		}\hspace{-7.5pt}
%		\hfill
%		\subfloat[Highlight]{
%			\begin{minipage}[b]{0.24\linewidth}
%				\centering
%				\includegraphics[width=\linewidth]{./pic5/3.png}\vspace{1pt}
%			\end{minipage}
%		}\hspace{-7.5pt}
%		\hfill
%		\subfloat[Non-highlight]{
%			\begin{minipage}[b]{0.24\linewidth}
%				\centering
%				\includegraphics[width=\linewidth]{./pic5/4.png}\vspace{1pt}
%			\end{minipage}
%		}\hspace{-7.5pt}
%		\hfill
%		
%	\end{minipage}
%	\vfill
%	\caption{Examples of the proposed RWHI dataset.}
%	\label{color11}\vspace{-0.4cm}
%\end{figure}
Our model is implemented in PyTorch on a GPU with NVIDIA GeForce 3090 and the input size of image is 224 $\times$ 224.
We use the Adam Optimizer to train our network with batch size of 16 and epochs of 100 which will require 12 hours.
In our experiments, we set the learning rate to $2\times 10^{-4}$ and then reduce it by half every 10 epochs.
\subsection{Comparison with Traditional Methods}

\begin{figure}[h]
	\centering		
	\subfloat[{Input \\ (PSNR/SSIM) }]{
	\includegraphics[width=0.18\textwidth,keepaspectratio]{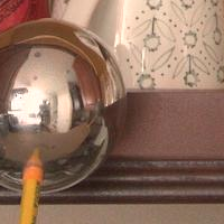}
	}
	\subfloat[Souza~\cite{Macedo2018}\\(22.35/0.79)]{
	\includegraphics[width=0.18\textwidth,keepaspectratio]{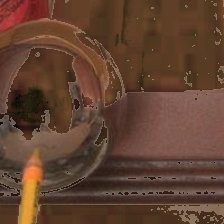}
	}
	\subfloat[Akashi~\cite{akashi2014separation}\\(27.08/0.61)]{
	\includegraphics[width=0.18\textwidth,keepaspectratio]{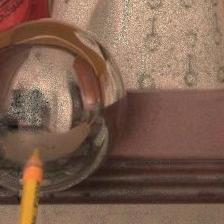}
	}
	\subfloat[Arnold~\cite{arnold2010automatic}\\(31.45/0.82)]{
	\includegraphics[width=0.18\textwidth,keepaspectratio]{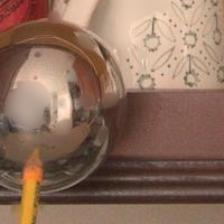}
	}
	\subfloat[{ours\\ (\textbf{\color{red}35.72/0.91})}]{
	\includegraphics[width=0.18\textwidth,keepaspectratio]{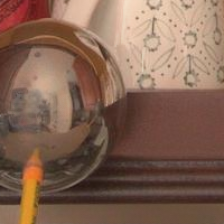}}
	\caption{Visual results of specular highlight removal and comparison with the traditional state-of-the-art methods~\cite{Macedo2018,akashi2014separation,arnold2010automatic}, and ours method on SHIQ~\cite{fu2021multi} dataset. }
	\label{traditional}
\end{figure}

% \begin{figure*}[htbp]
% 	\centering
% 	\subfloat[Title]{\includegraphics[width=.45\linewidth]{1.png}}\hspace{5pt}
% 	\subfloat[Title]{\includegraphics[width=.45\linewidth]{2.png}}\\
% 	\subfloat[Title]{\includegraphics[width=.45\linewidth]{3.png}}\hspace{5pt}
% 	\subfloat[Title]{\includegraphics[width=.45\linewidth]{4.png}}
% 	\caption{Description.}
% \end{figure*}

We compared the performance of our network with current state-of-the-art traditional methods~\cite{Macedo2018,akashi2014separation,arnold2010automatic}. As shown in the Fig.~\ref{traditional}, 
Souza~\cite{Macedo2018} can remove some of the highlights, and the metal surface is poorly removed with a large number of black patches. In contrast, the Akashi~\cite{akashi2014separation} and Arnold~\cite{arnold2010automatic} algorithms remove better results. However,  Arnold~\cite{arnold2010automatic} can not handle the boundary of the highlights and tend to preserve some weaker highlights. The highlight part is not natural with the non-highlight part. And the algorithm in this paper uses the CHA, which makes the filled pixels after highlight removal closer to the pixels in the non-highlight part, and the effect is better.
\begin{table}[h]
  \centering
  \caption{Quantitative comparison results between our method with previous methods on SHIQ~\cite{fu2021multi} and SD1~\cite{hou2021text} datasets. The evaluation metrics include PSNR, SSIM, and the best results are marked in {\color{red}red}.}
  	\renewcommand\tabcolsep{10pt}
	\renewcommand\arraystretch{1.1}
    \begin{tabular}{cccccc}
    \toprule
    \textbf{Datasets} & \multicolumn{2}{c}{\textbf{SHIQ}~\cite{fu2021multi}} &       & \multicolumn{2}{c}{\textbf{SD1}~\cite{hou2021text}} \\
\cmidrule{2-3}\cmidrule{5-6}    \textbf{Method/Metric} & \textbf{PSNR$\uparrow$} & \textbf{SSIM$\uparrow$} &       & \textbf{PSNR$\uparrow$} & \textbf{SSIM$\uparrow$} \\
    \midrule
    \textbf{Multi-class\cite{lin2019deep}} & \textbf{-} & \textbf{-} &       & \textbf{26.29} & \textbf{0.89} \\
    \textbf{SPEC\cite{muhammad2020spec})} & \textbf{19.56} & \textbf{0.69} &       & \textbf{15.61} & \textbf{0.69} \\
    \textbf{TA\cite{hou2021text}} & \textbf{-} & \textbf{-} &       & \textbf{22.65} & \textbf{0.88} \\
    \textbf{DeepFillv2\cite{yu2019free}} & \textbf{32.19} & \textbf{0.84} &       & \textbf{-} & \textbf{-} \\
    
    \textbf{JSHDR\cite{fu2021multi}} & \textbf{34.30} & \textbf{0.86} &       & \textbf{24.59} & \textbf{0.85} \\
    \hdashline
		\specialrule{0em}{1pt}{1pt}

    \textbf{Ours} & \textcolor[rgb]{1.000, 0.000, 0.000}{\textbf{35.72 }} & \textcolor[rgb]{1.000, 0.000, 0.000}{\textbf{0.91 }} &       & \textcolor[rgb]{1.000, 0.000, 0.000}{\textbf{33.44 }} & \textcolor[rgb]{1.000, 0.000, 0.000}{\textbf{0.92 }} \\	
	
    \bottomrule
    \end{tabular}%
  \label{tab:learning-based}%
\end{table}%

\subsection{Comparison with learning-based Methods}

%需要扬帆根据表格1补充这里的文字说明

% \noindent\textbf{Quantitative Evaluation.}
The results of quantitative evaluation on both SHIQ (Real-world highlight images)~\cite{fu2021multi} and SD1 (Text highlight images)~\cite{hou2021text} dataset are presented in Table~\ref{tab:learning-based}.
To effectively evaluate highlight removal performance, we adopt the metrics including PSNR and SSIM.
Through comparing the evaluation metrics of our method with existing methods, it can be observed that our proposed methods significantly outperforms other learning-based methods in terms of both metrics.
In specific, our method outperforms state-of-the-art highlight removal methods by more than 1.42 dB in PSNR and 0.05 in SSIM on the SHIQ dataset and 7.15 dB in PSNR and 0.07 in SSIM on the SD1 dataset.

\subsection{Comparison on Natural Datasets}
\begin{figure}[h]
	\centering
	\includegraphics[scale=0.4]{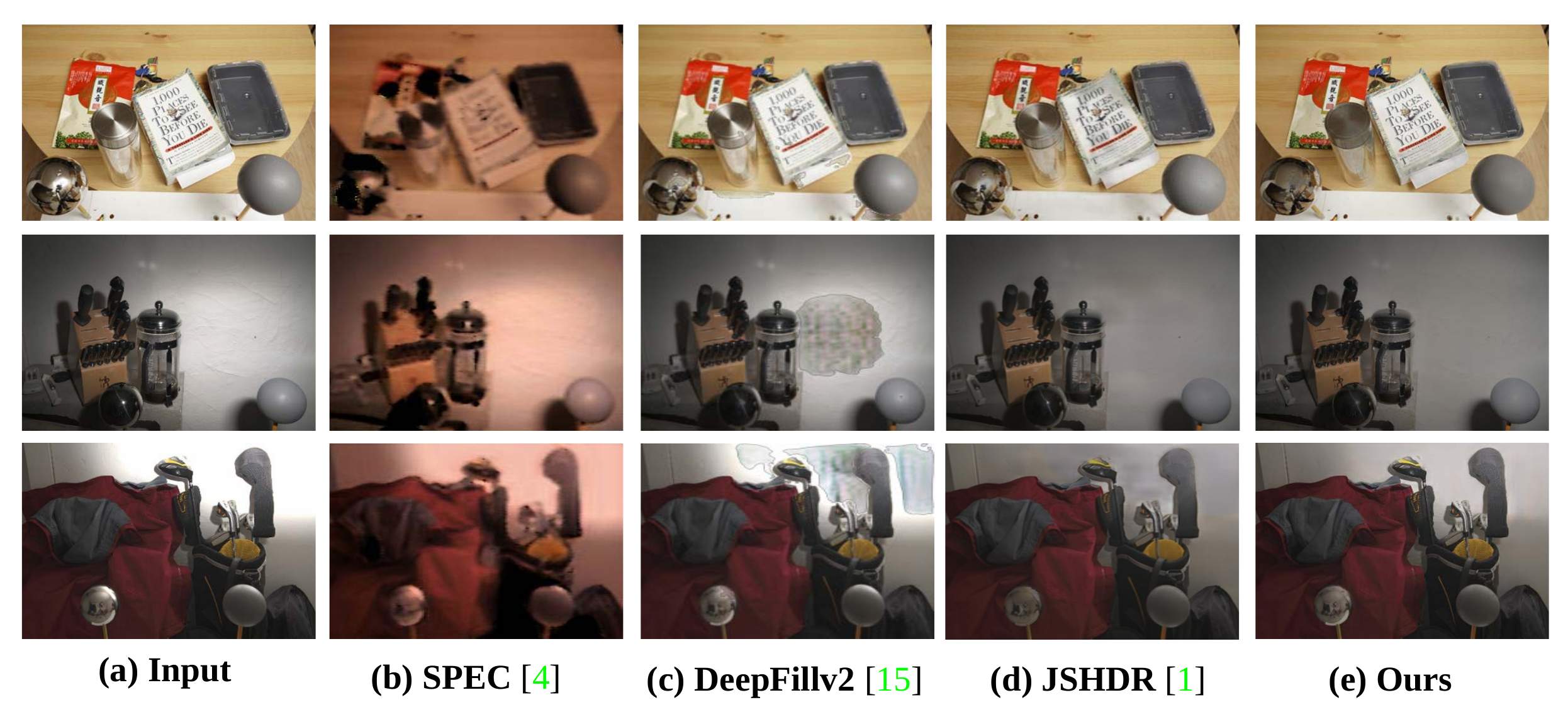}
	\caption{Compare with state-of-the-art algorithms (SPEC~\cite{muhammad2020spec}, DeepFillv2~\cite{yu2019free} and JSHDR~\cite{fu2021multi}) on natural images dataset~\cite{murmann19}.}
	\label{natural}
\end{figure}

% 图8展示了本文算法与现有最先进算法在自然图像数据集上的结果对比。三种不同光线照射的情况下，本文算法均能很好的处理金属表面的反光。与现有的最先进算法相比，本文的算法去除的高光图像的同时能很好的保持图像的完整性。并不会给高光去除的图像带来很明显的破坏，而算法DeepFillv2和JSHDR会在高光区域留下伪影，造成高光部分的像素与周围像素存在明显的差异。算法SPEC虽然不存在图像伪影的问题，但是会造成整张图像的颜色加深，视觉上与原始图像存在较大的差异性。
Figure~\ref{natural} shows the results of proposed algorithm compared with the existing state-of-the-art algorithms on the natural image dataset. The algorithms DeepFillv2~\cite{yu2019free} and JSHDR~\cite{fu2021multi} leave artifacts in the highlight area, causing significant differences between the pixels in the highlight section and the surrounding pixels. The algorithm SPEC~\cite{muhammad2020spec} does not have the problem of image artifacts, but it causes a deepening of the color of the whole image and a large visual difference from the original image. Compared with the existing state-of-the-art algorithms, the proposed algorithm removes the highlights while maintaining the integrity of the image and can handle the reflections on the metal surface well for all three different light illumination cases.
\vspace{-0.8cm}
\begin{figure}[H]
	\centering
	\includegraphics[scale=0.605]{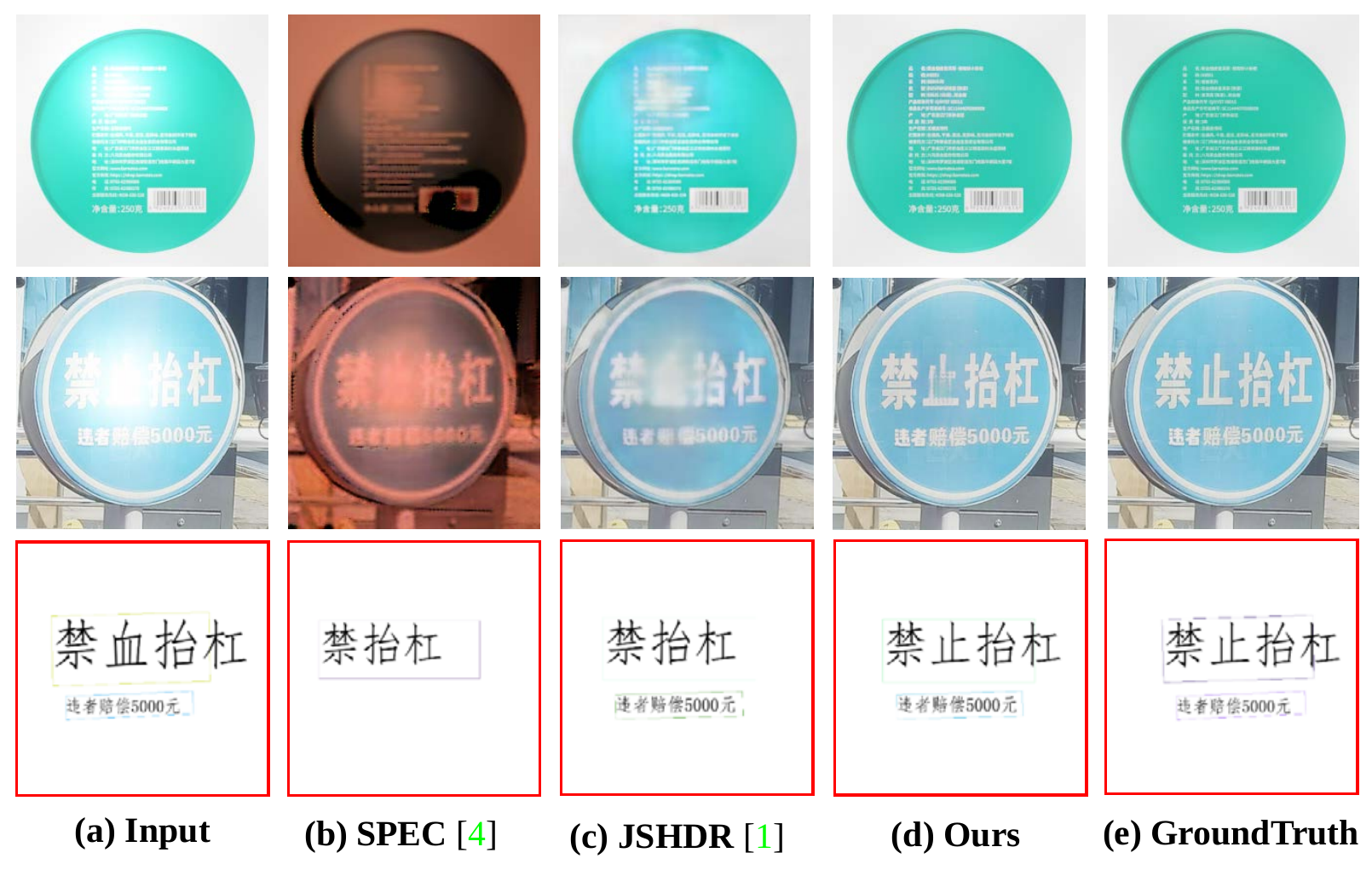}
	\caption{Compare with state-of-the-art algorithms on text images data. The first row show the comparative results of the different algorithms, and the last row shows the results of the recognition of the highlight removal image using the PaddleOCR (\url{https://github.com/PaddlePaddle/PaddleOCR}).}
	\label{text}

\end{figure}

\subsection{Comparison on Text Datasets}
% 可以看出，算法SPEC在文字图像上也会出现颜色加深的效果。高光污染的区域并没有得到实质性的效果提升。算法JSHDR的效果略微比算法SPEC的好一些，但是依然不能令人满意。我们的算法能很好的去除高光，在视觉上具有更加赏心悦目的效果。同时在不用OCR算法的辅助下也能肉眼识别出图片中的文字。
Fig.~\ref{text} shows the results of proposed algorithm compared with the existing state-of-the-art algorithms on the text image datasets. 
It can be seen that the algorithm SPEC~\cite{muhammad2020spec} also appears as a color deepening effect on text images. The areas with high light pollution do not get a substantial effect enhancement. The effect of algorithm JSHDR~\cite{fu2021multi} is slightly better than that of algorithm SPEC~\cite{muhammad2020spec}, but it is still unsatisfactory. Our algorithm removes highlights well and has a more visually pleasing effect. At the same time, the text in the image can be recognized by the naked eye without the assistance of the OCR algorithm.
\subsection{Comparison on Face Datasets}
\begin{figure}[h]
	\centering
	\includegraphics[scale=0.55]{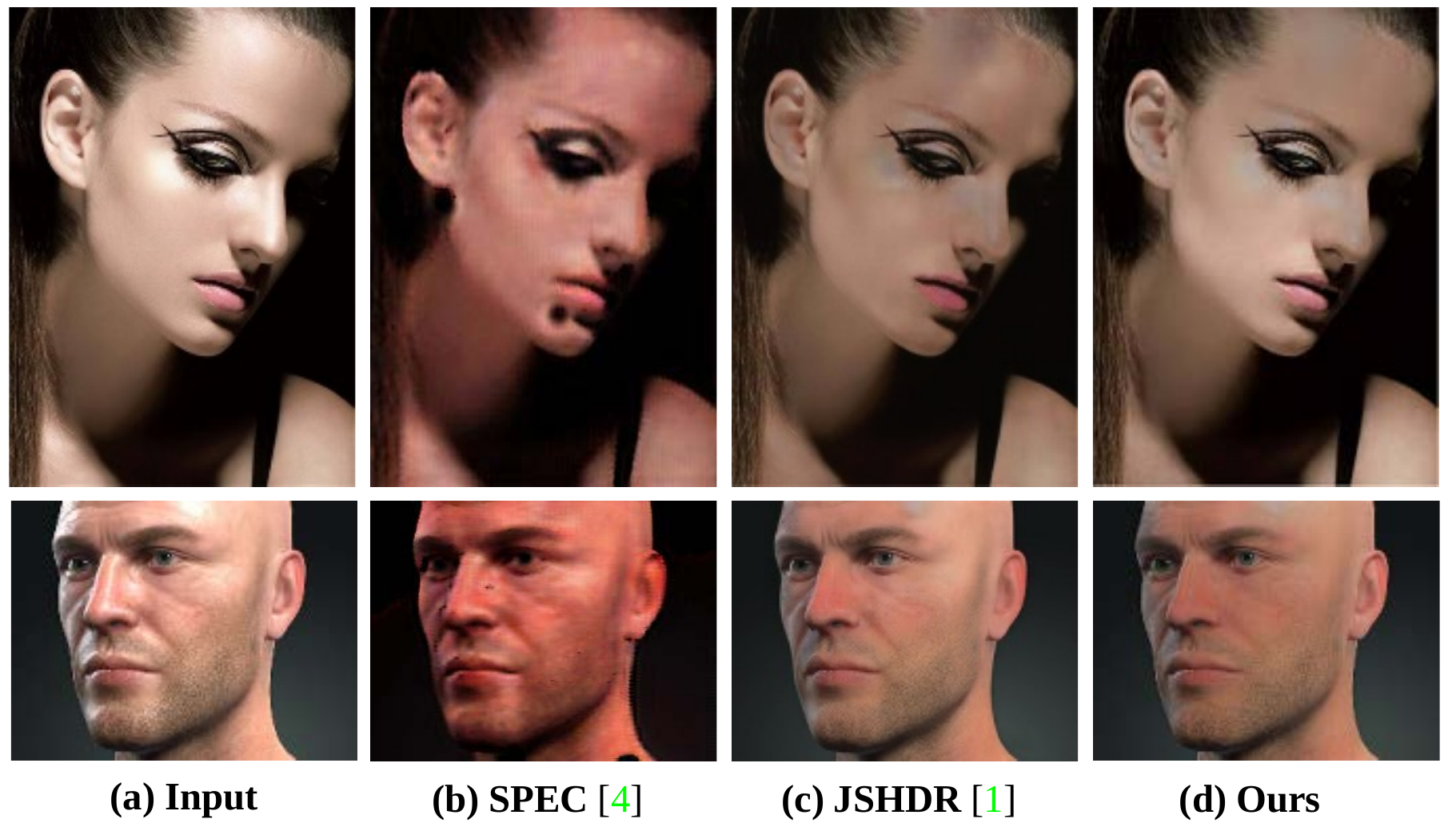}
	\caption{Compare with state-of-the-art algorithms (SPEC\cite{muhammad2020spec} and JSHDR~\cite{fu2021multi}) on face image data. The first row is a natural face image collected from the Internet and the second row is a synthetic face highlight image.}
% 	第一行为从互联网上收集的自然人脸图像，第二行为合成的人脸高光图像。
	\label{face}
\end{figure}

Fig.~\ref{face} shows the results of proposed algorithm compared with the existing state-of-the-art algorithms on the face dataset. It can be seen that the algorithm in this paper has the best result in removing highlights on both the real face data and the synthetic face data, and the removal effect is more natural. In contrast, SPEC~\cite{muhammad2020spec} and JSHDR~\cite{fu2021multi} algorithms will bring artifacts on the image while removing some highlights.  The SPEC~\cite{muhammad2020spec} algorithm also results in black circles in the chin area on natural images of the face.
\subsection{Results on Synthetic Image}

\begin{figure}[H]
	\centering
	\includegraphics[scale=0.35]{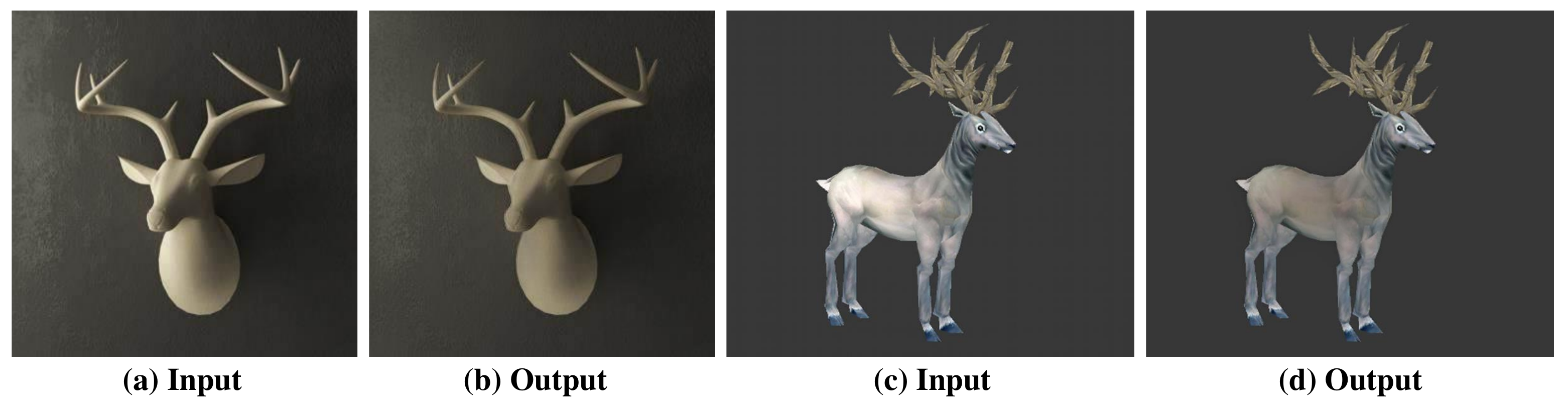}
	\caption{Highlight removal results of sythetic images. The first row is the synthetic highlight image and the second row is the output. The images were collected from the Internet.}
	\label{synthetic}
\end{figure}

Fig.~\ref{synthetic} shows the two sets of highlight images and the corresponding highlight removal images. It can be seen that the algorithm in this paper can achieve satisfactory results regardless of the highlight removal of small blocks (Fig.~\ref{synthetic} (a) and (b)) or large blocks (Fig.~\ref{synthetic} (c) and (d)).

\subsection{Application on Video Datasets}
\begin{figure}[h]
	\centering
	\includegraphics[scale=0.45]{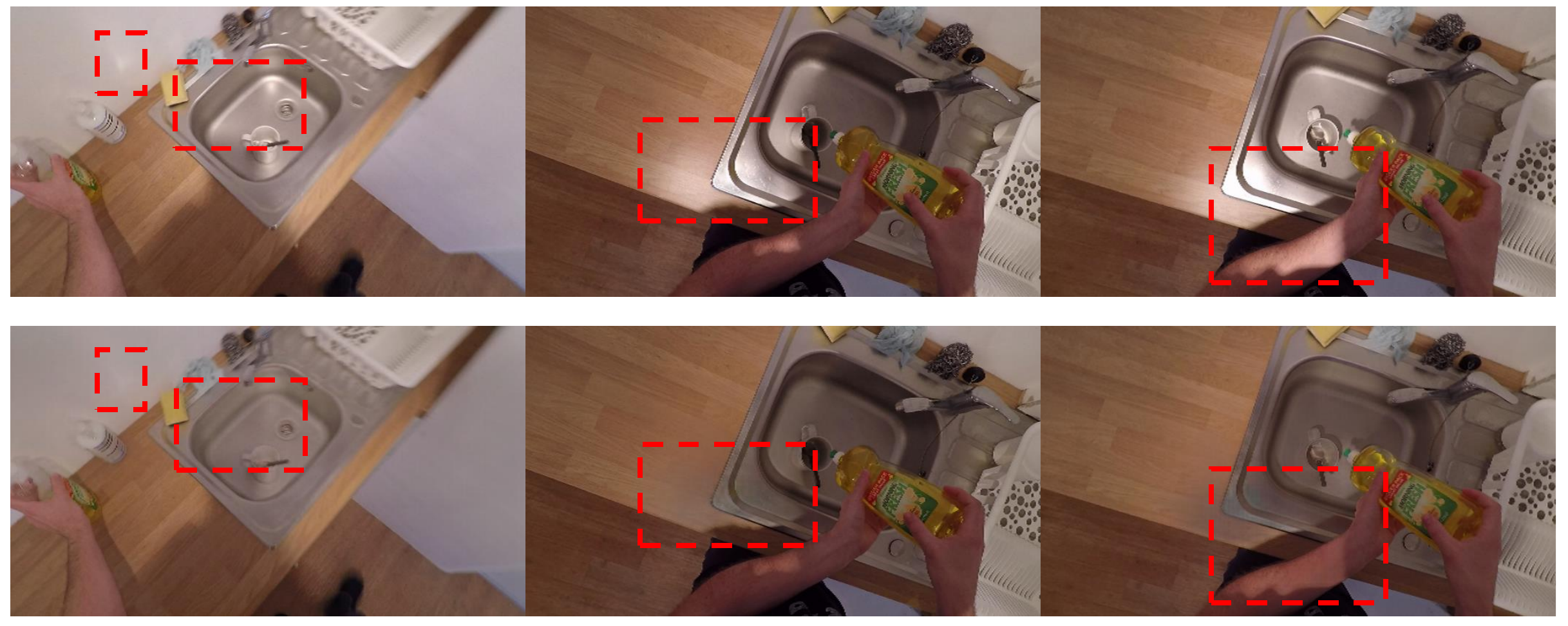}
	\caption{Ours algorithm is applied on highlight video. The video resolution selected is $1920 \times 1080$. A frame for every 0.5 seconds. Red dashed line represents the contrast of the main highlight region.}
	\label{vid}
\end{figure}

As far as we know, there is no algorithm to remove video highlights. In order to demonstrate the superiority of the proposed algorithm, we use the image highlight removal our algorithm directly for the video highlight processing. Specifically, the video is exported on a per-frame basis, and then processed for each image frame of the video. Fig.~\ref{vid} shows the video highlight removal effect. We can see that the highlight regions are well removed, and many small highlight regions are also precisely removed.
% 据我们所知，目前还没有去除视频高光的算法。为了体现本文算法的优越性，我们将本文图像高光去除算法直接用于视频的高光处理。具体地，将视频按照每帧导出，然后针对视频的每帧图像进行处理。
\vspace{-0.3cm}

\section{Discussion}

\subsection{Ablation Study}
To verify the effectiveness of the proposed components of our network, we perform the ablation experiment by modifying the components.
\begin{table}[h]
  \centering
  \caption{Quantitative comparison results of ablation study for specular highlight removal on SHIQ~\cite{fu2021multi} and SD1~\cite{hou2021text} datasets. The best and second best results are marked in {\color{red}red} and {\color{blue}blue}, respectively. Baseline: only Coarse Network and Refine Network, excluding our proposed
HFE and CHA. DenseUnet: use only DenseUnet to remove highlight directly. CHA consists of BA and HA.}
  	\renewcommand\tabcolsep{10pt}
	\renewcommand\arraystretch{1.1}
    \begin{tabular}{cccrcc}
    \toprule
    \textbf{Datasets} & \multicolumn{2}{c}{\textbf{SHIQ}~\cite{fu2021multi}} &       & \multicolumn{2}{c}{\textbf{SD1}~\cite{hou2021text}} \\
\cmidrule{2-3}\cmidrule{5-6}    \textbf{Method/Metric} & \textbf{PSNR$\uparrow$} & \textbf{SSIM$\uparrow$} &       & \textbf{PSNR$\uparrow$} & \textbf{SSIM$\uparrow$} \\
    \midrule
    \textbf{Baseline} & \textbf{31.08 } & \textbf{0.86 } &       & \textbf{23.35 } & \textbf{0.82 } \\
    		\hdashline
		\specialrule{0em}{1pt}{1pt}
    \textbf{w/ HFE  w/o CHA} & \textbf{32.64 } & \textbf{0.84 } &       & \textbf{25.67 } & \textbf{0.87 } \\
    \textbf{w/ HFE\&BA} & \textbf{31.71 } & \textbf{0.87 } &       & \textbf{28.46 } & \textbf{0.89 } \\
    \textbf{w/ HFE\&HA} & \textcolor[rgb]{0.000, 0.439, 0.753}{\textbf{33.79 }} & \textcolor[rgb]{0.000, 0.439, 0.753}{\textbf{0.89 }} &       & \textcolor[rgb]{0.000, 0.439, 0.753}{\textbf{33.06 }} & \textcolor[rgb]{0.000, 0.439, 0.753}{\textbf{0.90 }} \\
    \textbf{DenseUnet} & \textbf{29.18 } & \textbf{0.83 } &       & \textbf{22.38 } & \textbf{0.84 } \\
    		\hdashline
% 		\specialrule{0em}{1pt}{1pt}
    \textbf{Ours} & \textcolor[rgb]{1.000, 0.000, 0.000}{\textbf{35.72 }} & \textcolor[rgb]{1.000, 0.000, 0.000}{\textbf{0.91 }} &       & \textcolor[rgb]{1.000, 0.000, 0.000}{\textbf{33.44 }} & \textcolor[rgb]{1.000, 0.000, 0.000}{\textbf{0.92 }} \\
    \bottomrule
    \end{tabular}%
  \label{tab:ablation}%
\end{table}%
% \textbf{\color{red}To verify the effectiveness of the proposed components of our network, we perform the ablation experiment by modifying the components.
% First, to test the effectiveness of our HFE module, we made two variations on it.
% One option is to remove the HFE module directly and replace it with a unit tensor, which implies that both the highlight region and non-highlight region will be generated by the network.
% The other option is to use a CNN-based network in place of the HFE module.
% We denote these two changes as "w/o HFE" and "w/ Convs" respectively.}
% These variants can be described by the following:
% % First,we set a baseline that includes only Coarse Network and Refine Network, excluding our proposed HFE and CHA, which we denote as "Baseline".
% % Then we add the HFE to test the 
% \begin{itemize}
% 	\item Baseline: only Coarse Network and Refine Network, excluding our proposed HFE and CHA.
% 	\item w/ HFE w/o CHA: add HFE module to "Baseline".
% 	\item w/ HFE\&BA: add HFE and BA to "Baseline".
% 	\item w/ HFE\&HA: add HFE and HA to "Baseline".
% 	\item DenseUnet: use only DenseUnet to remove highlight directly.
% \end{itemize}
% to test the effectiveness of our HFE module, we made two variations on it.
% One option is to remove the HFE module directly and replace it with a unit tensor, which implies that both the highlight region and non-highlight region will be generated by the network.
% The other option is to use a CNN-based network in place of the HFE module.
% We denote these two changes as "w/o HFE" and "w/ Convs" respectively.
\begin{figure}[h]
	\centering
	\includegraphics[scale=0.4]{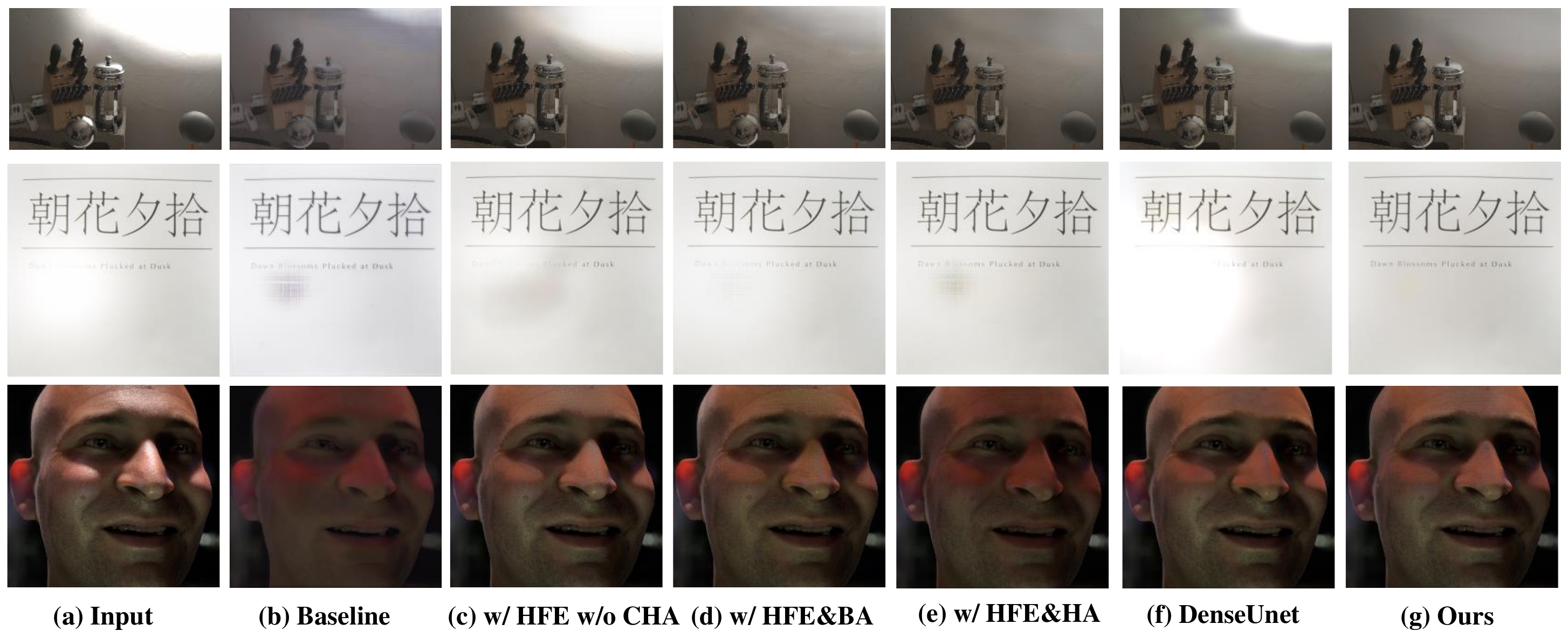}
	\caption{Ablation results for our proposed components. The first row is the natural highlight image, the second row is the text highlight image, and the third row is the face highlight image.}
	\label{Ablation}
\end{figure}
Table~\ref{tab:ablation} show the PSNR and SSIM of the ablation study and Fig.~\ref{Ablation} show the visual results. To be fair, we use the same dataset to train the network.

From the results, we can draw the following conclusions:
(1) Our proposed HFE module plays an essential role in the task of highlight removal. 
Without HFE module, it is impossible for the generator to produce a satisfactory result.
(2) The CHA module can improve the performance of highlight removal.
As shown in Fig.~\ref{Ablation} (c), without the CHA, the network may not ensure the colors and can't address the large scale highlight region.
(3) From table and Fig.~\ref{Ablation} (d-e), it can be observed that BA tends to keep more background information, HA tends to keep more highlight  information. The two attention work together to ensure the best performance.
(4) The poor results of highlight removal directly using DenseUnet prove the validity of our model, not for fitting on just the training dataset.

\subsection{Limitation}
\begin{figure}[hbt]
	\centering
	\includegraphics[scale=0.16]{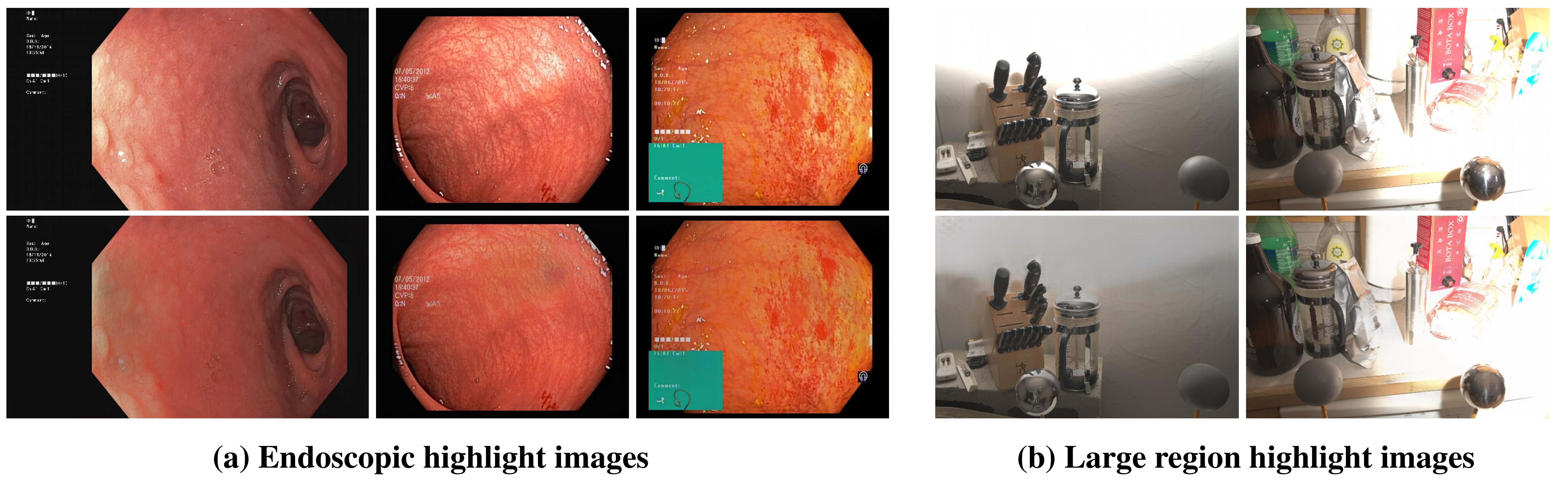}
	\caption{Scenarios and cases where the proposed algorithm does not work well. (a) Comparison of the removal effect of three endoscopic highlight images; (b) Comparison of the removal effect of extra-large region highlight under natural scene.}
	\label{limitation}
\end{figure}

% 这里展示内窥镜的效果，效果不好没有关系
Although our proposed method has achieved an excellent result in highlight removal, it does have some limitations.
For unsaturated highlights, our model can effectively remove them, However, for saturated highlights, our model can not recover reliable textures. The effect of our algorithm on the endoscopic images is shown in Fig.~\ref{limitation} (a). It can be seen that for the large red endoscope dataset, the proposed algorithm does not measure the highlight removal well and produces patches that are very different from the surrounding cells. Besides, for extra-large region highlight images removal, our proposed algorithm is not yet able to handle it well in Fig.~\ref{limitation} (b). It is mainly because the extra-large highlight region severely affects the contextual attention mechanism and can not fill the large region of missing pixels with non-highlight regions.

% 本文的算法在内窥镜上的效果见图3.可以看出，针对大面积的红色内窥镜数据集，本文的算法并不能很好测去除高光部分，而且会产生与周围细胞差异性很大的斑块。

\section{Conclusion}

% 我们提出了一个端到端的多场景的多阶段的高光检测和高光去除网络。特别的，高光检测网络运用多尺度图像信息实现了高光区域的准确检测。结合高光信息和背景信息的上下文注意力机制的使得架构能去除高光的同时实现与周围像素更加接近的填补效果。通过大量的实验结果揭示了提出的算法在合成图像、文字图像、人脸图像和自然图像上与现有最先进的算法相比都取得了更好的结果。特别的，本文的算法首次在视频高光去除上也取得了令人满意的效果。
We propose a multi-scenes and multi-stages highlight detection and highlight removal network. In particular, the highlight extractor uses multi-scale image information to achieve accurate detection of highlight regions. The contextual highlight attention mechanism that combines highlight information and background information enables the architecture to remove highlights while achieving a closer fill effect to the surrounding pixels. Extensive experimental results reveal that the proposed algorithm achieves better results on synthetic images, text images, face images, and natural images compared to the existing state-of-the-art algorithms. In particular, the proposed algorithm achieves satisfactory results on video highlight removal for the first time as well.

% In this paper, we proposed a novel highlight removal model and network specifically for highlight removal.
% Compared with existing models, it achieves a higher level of effectiveness in highlight removal.
% In this network, no longer need highlight mask, the network is trained on pairs of highlight images and non-highlight images, which alleviates the demand for high-quality to a certain extent.
% Quantitative results, qualitative comparisons, and ablation studies have demonstrated the superiority of the proposed highlight removal model and network.
% In addition, a real-world highlight dataset is presented contribute to improve the removal performance in the real world.

% In the future work, we intend to extend our model to other computer vision tasks such as reflection removal \cite{lei2021robust}, moire pattern removal \cite{he2019mop}, and shadow removal.

%References are listed in alphabetic order by the surname of the first author, or the identifying word (e.g., in case of a website). Have
%all anonymized references at the beginning of the list.

%here would be your acknowledgement (if any) in the final accepted paper

%===========================================================
\bibliographystyle{splncs}
\bibliography{egbib}

%this would normally be the end of your paper, but you may also have an appendix
%within the given limit of number of pages
\end{document}

\pagestyle{headings}
\mainmatter
\author{Anonymous ACCV 2022 submission}

\def\ACCV22SubNumber{200}  % Insert your submission number here

%===========================================================
\title{$M^2$-Net:Multi-stages Specular Highlight Detection and Removal in Multi-scenes Supplementary Materials} % Replace with your title
\titlerunning{ACCV-22 submission ID \ACCV22SubNumber}
\authorrunning{ACCV-22 submission ID \ACCV22SubNumber}

% \author{Anonymous ACCV 2022 submission}
\institute{Paper ID \ACCV22SubNumber}

\maketitle

\vspace{-0.5cm}
\begin{figure}[ht]
	\centering
	\includegraphics[scale=0.25]{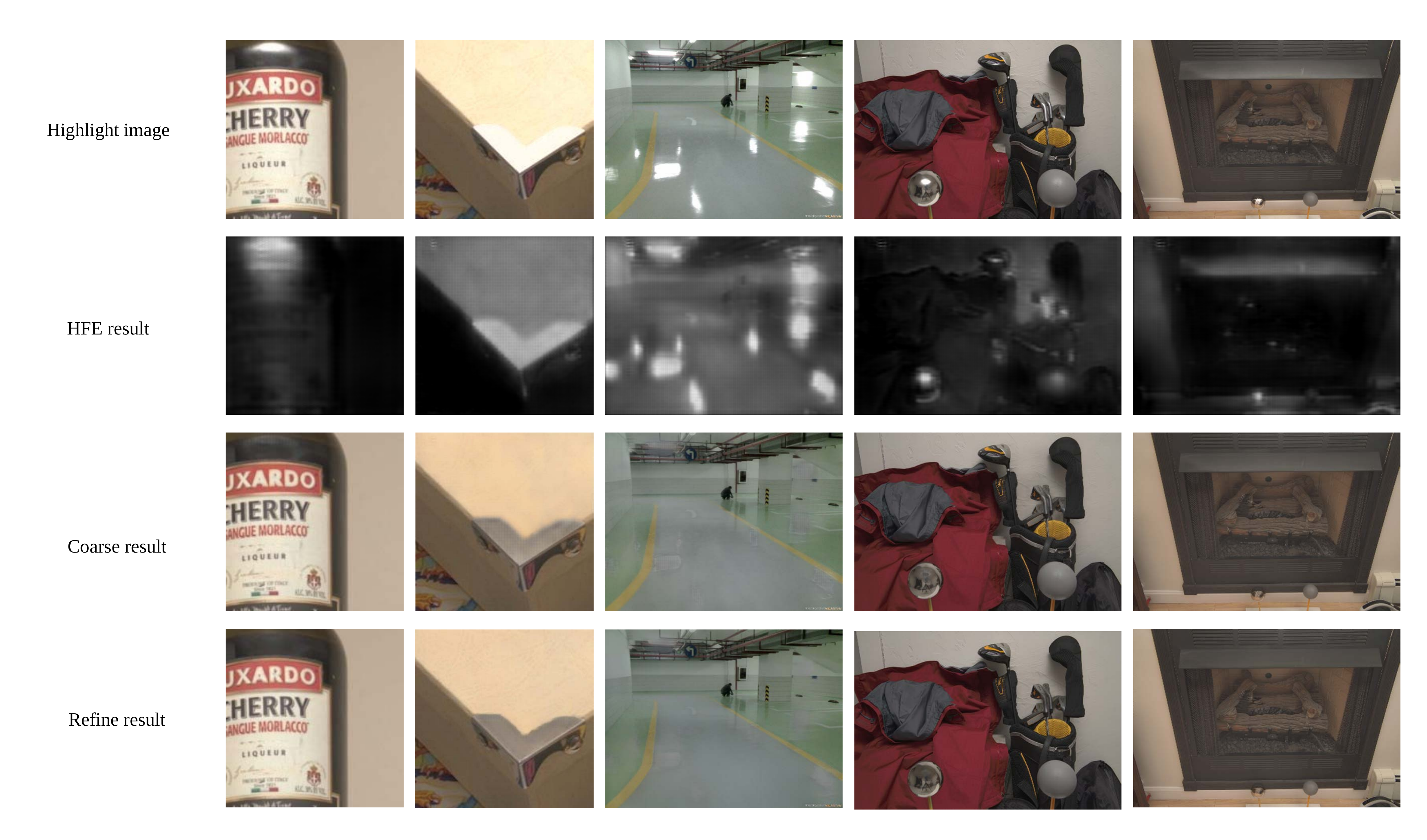}
	\caption{ The results of our $M^2$-Net .}

	\label{111}
\end{figure}
\vspace{-0.8cm}

\section{Experiment Details}
\label{Details}

Our network is implemented in PyTorch on a GPU(NVIDIA GeForce 3090)  and the input size of image is 224 $\times$ 224.
We use Adam Optimizer to train the network with the batch size of 16 and the epochs of 300 which takes around 1 day.
In the experiment, we set an initial learning rate of $2 \times 10^{-4}$ and adjust it by half every 10 epochs.

To train the network more efficiently, we use the SHIQ\cite{fu2021multi} and SD1\cite{hou2021text} as the training dataset, and test the network on SHIQ, SD1 and Multi-Illumination Images datasets\cite{murmann19}.

\section{$M^2$-Net Examples}
See Figure~\ref{111} for additional examples of  steganography quality results on the images from internet. 
As show in the figure, the first row is the highlight images and the second row is the highlight detection results by out HFE module, then the third and fourth row are Coarse results and Refine results produced by the coarse highlight removal network and refine highlight removal network respectively.

\section{More Facial Highlight Removal Examples}

\begin{figure}[h]
	\centering
	\includegraphics[scale=0.4]{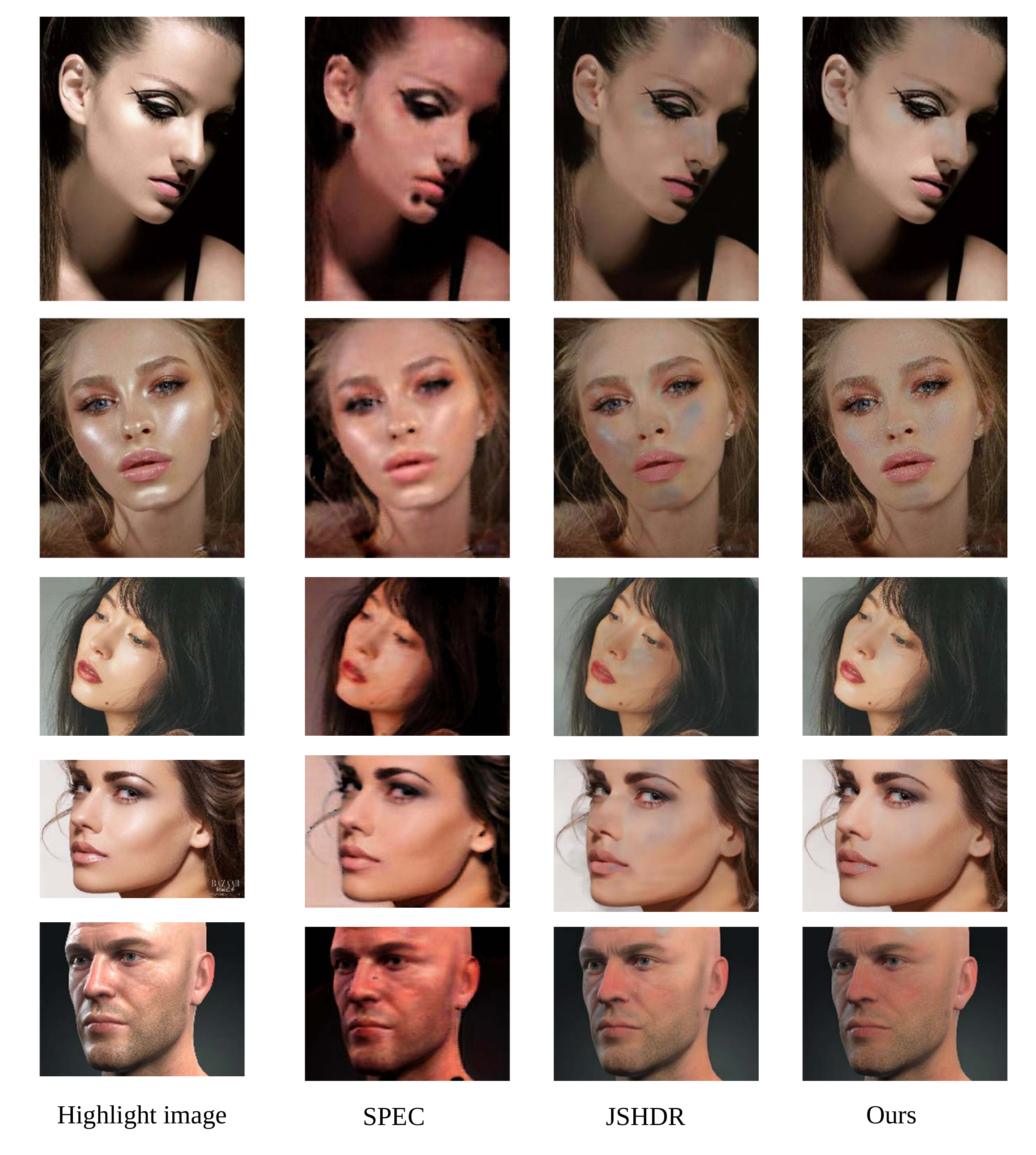}
	\caption{ The results of facial highlight removal .}

	\label{222}
\end{figure}

As show in the Figure~\ref{222}, We show more results of facial highlight removal with images from the Internet or other datasets.

\vspace{-0.5cm}
\vspace{-0.8cm}
\section{More Text Highlight Removal Examples}

\begin{figure}[h]
	\centering
	\includegraphics[scale=0.3]{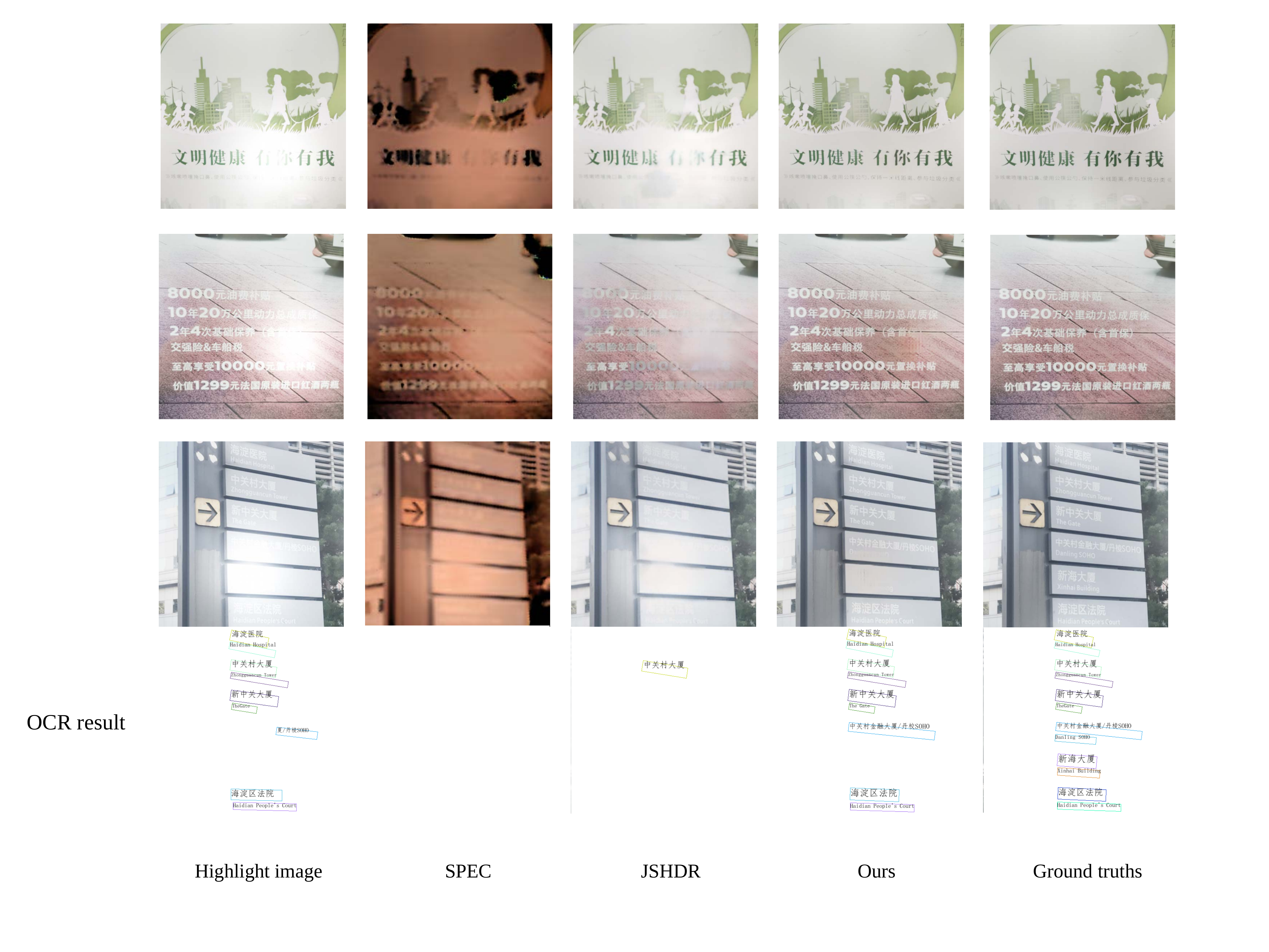}
	\caption{ The results of text highlight removal .}

	\label{333}
\end{figure}

As show in the Figure~\ref{333}, We show more results of text highlight removal and OCR  with images from the Internet or other datasets.

\section{Code}
\label{sec:code}
Our code and the trained model will be placed at  \url{https://github.com/hzzzyf/specular-removal}

\vspace{-0.5cm}
\vspace{-0.8cm}
%===========================================================
\bibliographystyle{splncs}
\bibliography{egbib}

%this would normally be the end of your paper, but you may also have an appendix
%within the given limit of number of pages